\documentclass[11pt]{article}
\usepackage[margin=1in]{geometry}
\usepackage{amsmath,amssymb}
\usepackage{booktabs}
\usepackage{graphicx}
\graphicspath{{./}{./figures/}{./images/}}
\usepackage{siunitx}
\usepackage{hyperref}
\usepackage[authoryear]{natbib}
\usepackage{microtype}
\usepackage{authblk}
\usepackage{enumitem}
\usepackage{titlesec}
\usepackage{caption}
\usepackage{setspace}
\usepackage{longtable}
\usepackage{algorithm}
\usepackage{amsmath}
\usepackage{algpseudocode}
\usepackage{amsfonts}
\usepackage{float}

\titleformat{\section}{\large\bfseries}{\thesection}{0.75em}{}
\titleformat{\subsection}{\normalsize\bfseries}{\thesubsection}{0.75em}{}
\titleformat{\subsubsection}{\normalsize\itshape}{\thesubsubsection}{0.75em}{}

\title{Enhancing Transformer-Based Foundation Models for Time Series Forecasting via Bagging, Boosting and Statistical Ensembles}
\author[1]{Dhruv D. Modi}
\author[1]{Rong Pan}
\affil[1]{School of Computing and Augmented Intelligence, Arizona State University}
\date{\today}

\begin{document}
\maketitle

\begin{abstract}
Time series foundation models (TSFMs) such as Lag-Llama, TimeGPT, Chronos, MOMENT, UniTS, and TimesFM have shown strong generalization and zero-shot capabilities for time series forecasting, anomaly detection, classification, and imputation. Despite these advantages, their predictions still suffer from variance, domain-specific bias, and limited uncertainty quantification when deployed on real operational data. This paper investigates a suite of statistical and ensemble-based enhancement techniques---bootstrap-based bagging, regression-based stacking, prediction-interval construction, statistical residual modeling, and iterative error feedback---to improve robustness and accuracy. Using the Belgium Electricity Short-Term Load Forecasting dataset as a case study, we demonstrate that the proposed hybrids consistently outperform standalone foundation models across multiple horizons. Regression-based ensembles achieve the lowest mean squared error; bootstrap aggregation markedly reduces long-context errors; residual modeling corrects systematic bias; and the resulting prediction intervals achieve near-nominal coverage with widths shrinking as context length increases. The results indicate that integrating statistical reasoning with modern foundation models yields measurable gains in accuracy, reliability, and interpretability for real-world time series applications.
\end{abstract}

\noindent\textbf{Keywords:} time series forecasting; foundation models; Transformers; ensembling; bagging; boosting; uncertainty quantification

\section{Introduction}
Accurate time series forecasting plays a central role in decision-making for energy, finance, healthcare, and retail. Traditional statistical models, including ARIMA/SARIMA and exponential smoothing, remain widely used because they are interpretable and computationally light \citep{montgomery2015introduction, liang2024foundation}. However, they often struggle with non-linear dependencies, regime changes, and long-range structure. Deep learning approaches based on CNNs and RNNs/LSTMs mitigate some of these issues, yet face limitations in capturing very long dependencies and often require heavy task-specific tuning.

Transformer architectures address these issues with self-attention and scale favorably with data \citep{jiang2024empowering}. Recently, \emph{foundation models} for time series have emerged, leveraging pretraining and adaptation to deliver zero-shot generalization and strong accuracy across tasks and domains \citep{rasul2023lagllama, garza2023timegpt, ansari2024chronos, goswami2402moment, gao2024units, woo2402unified}. Still, three obstacles remain persistent in practice: (i) high variance for single-model forecasts, (ii) domain-specific bias when pretrained models meet specialized seasonal or calendar effects, and (iii) insufficient uncertainty quantification when only point forecasts are provided.

This work augments foundation models with statistical post-processing and hybridization. We propose: (1) bootstrap-based bagging to stabilize probabilistic forecasts; (2) regression-based ensembling that balances complementary inductive biases; (3) principled prediction intervals (PIs) for uncertainty communication; and (4) residual modeling and iterative error feedback for bias correction and online adaptation. We evaluate these methods on an operational electricity load forecasting task and show consistent improvements over foundation-model baselines. This paper represents a condensed version of Dhruv Modi's master's thesis, \emph{Foundation Models for Time Series}, completed at Arizona State University in 2025 \citep{modi2025foundation}.  

\textbf{Contributions.} We provide a unified methodology for integrating bagging, stacking, prediction intervals (PIs), and residual boosting with Transformer-based foundation models; present empirical evidence that these hybrids improve 1--24 hour forecasting accuracy and reduce variance on a real-world load dataset; and deliver a PI construction for ensembles that achieves near-nominal coverage while narrowing with longer historical context.

The rest of this paper is organized as follows: Section 2 reviews related work in statistical and foundation model-based time series forecasting. Section 3 details the proposed methodology, including bagging, boosting, and ensembling strategies. Section 4 describes the experimental setup. Section 5 presents results and discussion, and Section 6 concludes with implications and directions for future research.

\section{Related Work}
\subsection{Classical Forecasting}
Autoregressive and moving-average models, exponential smoothing, and their seasonal variants remain cornerstones of industrial forecasting \citep{montgomery2015introduction, liang2024foundation}. Their strengths include interpretability and strong performance on stationary or near-stationary processes, yet they can underfit complex nonlinear structure and long-range dependencies.

\subsection{Deep Learning for Time Series}
CNNs and RNNs/LSTMs have been adapted to forecasting, often improving non-linear modeling. Nevertheless, they may suffer from vanishing gradients, limited receptive fields, and heavy hyperparameter tuning for long-context tasks.

\subsection{Transformer-Based Foundation Models}
Self-attention enables modeling of long-range dependencies without recurrence \citep{jiang2024empowering}. Several time-series foundation models (TSFMs) extend this paradigm: \textbf{Lag-Llama} is a decoder-only univariate forecaster using RMSNorm and RoPE with a distribution head for probabilistic outputs \citep{rasul2023lagllama, liang2024foundation, liang2024foundation}. \textbf{TimeGPT} adds multi-horizon zero-shot forecasting with exogenous variables in an encoder--decoder design \citep{garza2023timegpt}. \textbf{Chronos} treats time series as tokens via scaling and quantization to repurpose T5-like language models for forecasting and quantiles \citep{ansari2024chronos, jiang2024empowering}. \textbf{MOMENT} uses masked time-series modeling with an encoder-only backbone and patching \citep{goswami2402moment}. \textbf{UniTS} introduces unified multitask and multidomain modeling via task tokenization and dynamic attention \citep{gao2024units}. \textbf{TimesFM} employs patch-based tokenization and decoder-only transformers trained on a mixture of real and synthetic data for long-horizon forecasting \citep{woo2402unified}. While these works demonstrate strong generalization, systematic study of statistical hybridization (bagging, stacking, residual correction) for TSFMs remains limited.

\section{Methodology}
We integrate bootstrap-based bagging, regression-based stacking, prediction-interval construction, residual modeling, and iterative error feedback into forecasting pipelines built on foundation models. The methods target variance reduction, bias correction, and uncertainty reporting. We use AutoGluon \citep{shchur2023autogluon}, which is an open-source AutoML toolkit from AWS, to automate statistical/machine-learning model selection, ensembling, and hyperparameter tuning. 

\subsection{Bagging: Bootstrap-Based Enhancement}
For each time step, the base foundation model (Lag-Llama) produces a point forecast and a sample of $n{=}100$ one-step-ahead draws $\{\hat{y}^{(i)}_t\}_{i=1}^n$. We draw $b{=}40$ values with replacement to compute a bootstrap mean $\bar{y}^{(j)}_t = \frac{1}{b}\sum_{i=1}^b \hat{y}^{(i)}_t$, repeat this $m{=}100$ times, and average to obtain the final forecast $\hat{y}^{\text{bag}}_t = \frac{1}{m}\sum_{j=1}^m \bar{y}^{(j)}_t$. This aggregation reduces variance relative to single-sample predictions.

\begin{algorithm}[H]
\caption{Bootstrap-Enhanced Lag-Llama Forecasting}\label{alg:bootstrapped-lagllama}
\begin{algorithmic}[1]
\Statex\textbf{Input}: Univariate time series
\Statex\textbf{Output}: Prediction from Point Prediction Model and the Bagging model along with their MSE.
\Statex \textbf{Step a: Lag-Llama Prediction Process:}
    \State Select past \{1 week, 3 weeks, 5 weeks\} as context window. 
    \State Perform zero shot forecasting to obtain \textbf{Point Prediction} and \textbf{sample of 100 values} for next hour (one-step-ahead) forecast.
    \State The context window slides by 1 hour, and the process repeats.
\Statex  \textbf{Step b: Bootstrap Mean Calculation:} 
    \State Sample 40 values from the 100 values (with replacement) and compute the mean.
\Statex  \textbf{Step c: Generate 100 Bootstrap Mean Columns:} 
    \State Apply Step b to the prediction sample 100 times.
\Statex \textbf{Step d: Compute Mean:}
        \State bootstrap\_mean\_all: Mean of all bootstrap mean columns generated in Step c.
\Statex  \textbf{Step e: Evaluation:}
    \State Compute MSE between:
    \begin{itemize}
        \item Point Prediction and actual.
        \item Compute MSE between bootstrap\_mean\_all and actual.
    \end{itemize}
\end{algorithmic}
\end{algorithm}

Algorithm \ref{alg:bootstrapped-lagllama}: Bootstrap-Enhanced Lag-Llama Forecasting improves the accuracy of univariate time series predictions by integrating statistical bagging techniques with the transformer-based Lag-Llama model. The process begins with Step a, where prediction from Lag-llama is obtained. Here, model selects a contextual window of historical data (e.g., 1 week, 3 weeks, or 5 weeks) and performs zero-shot forecasting to generate a Point Prediction along with a sample of 100 possible next-hour forecasts. This procedure is applied iteratively as the context window slides forward by one hour. Step b introduces a bootstrap-based function, where 40 values are randomly sampled (with replacement) from the 100 forecasted values, and their mean is computed. This function is invoked 100 times in Step c, generating 100 different bootstrap mean values, effectively creating an ensemble of forecasts. In Step d, the final prediction is obtained by computing the mean of all bootstrap means, producing a more robust and stable estimate. Lastly, Step e evaluates the model's accuracy by computing the Mean Squared Error (MSE) for both the Point Prediction and the bootstrap-enhanced prediction against the actual observed values. By incorporating bagging (bootstrap aggregation), this approach aims to mitigate prediction variance and enhance the reliability of Lag-Llama forecasts, demonstrating the effectiveness of statistical ensembling techniques in transformer-based time series forecasting.

\subsection{Regression Stacking of Lag-Llama and AutoGluon}
We form a linear stack of Lag-Llama and AutoGluon predictions,
\begin{equation}
\hat{y}^{\text{ens}}_t = w_1 \hat{y}^{\text{Lag}}_t + w_2 \hat{y}^{\text{AG}}_t,
\end{equation}
where $(w_1,w_2)$ are fit via time-series cross-validated linear regression on the training set. The combination exploits complementary strengths: long-range dependency modeling from the Transformer and strong local/seasonal structure from statistical/ML ensembles.

\begin{algorithm}
\caption{Ensemble-Based Model for Lag-Llama and AutoGluon Predictions}
\label{alg:ensemble_model}
\begin{algorithmic}[1]
\Statex \textbf{Input}: Univariate time series data.
\Statex \textbf{Output}: Optimal regression model with best combination of predictions.

\Statex \textbf{Step a: Generate Predictions from Lag-Llama and AutoGluon}
    \State Select past \{1 week, 3 weeks, 5 weeks\} as context window and generate 24-step-ahead forecasts using both Lag-Llama and AutoGluon.
    \State The context window \textbf{slides by 1 hour}, and the process repeats.
\Statex \textbf{Step b: Build Ensemble Model for all Step ahead Forecasts}
\Statex \textbf{(i) Simple Averaging}: Compute the average of Lag-Llama and AutoGluon predictions:
\[
avg = \frac{step_x\_lagllama + step_x\_autogluon}{2}
\]

\Statex \textbf{(ii) Weighted Combinations}: Compute weighted averages with different weight distributions:
\[
0.4L + 0.6A = 0.4 \cdot step_x\_lagllama + 0.6 \cdot step_x\_autogluon
\]
\[
0.25L + 0.75A = 0.25 \cdot step_x\_lagllama + 0.75 \cdot step_x\_autogluon
\]
\[
0.75L + 0.25A = 0.75 \cdot step_x\_lagllama + 0.25 \cdot step_x\_autogluon
\]

\Statex \textbf{(iii) Regression-Based Ensemble}: Fit a regression to get optimal weights.
\State Split data into training (80\%) and testing (20\%), maintaining time order.
\State Train a Linear Regression model using Lag-Llama and AutoGluon predictions as input features by performing Time Series Cross-Validation with 5 splits on training data.
\[
regression = w_{1x} \cdot step_x\_lagllama + w_{2x} \cdot step_x\_autogluon
\]

\Statex \textbf{Step c: Evaluate}
\State Calculate the MSE of Individual and Ensemble Models w.r.t actual value.

\end{algorithmic}
\end{algorithm}

Explanation of Notations used in algorithm \ref{alg:ensemble_model}:
\begin{itemize}
    \item $step_x\_lagllama$: The forecasted value for step-x-ahead (i.e., x hours ahead) generated by the Lag-Llama model.
    
    \item $step_x\_autogluon$: The forecasted value for step-x-ahead generated by the AutoGluon model.
    
    \item $w_{1x}$, $w_{2x}$: The weights learned by the Linear Regression model for Lag-llama and Autogluon respectively, that determine how much influence each model's prediction has in the final ensemble output for step-x-ahead.
\end{itemize}

Algorithm \ref{alg:ensemble_model}: Ensemble-Based Model for Lag-Llama and AutoGluon Predictions, combines forecasts from two different models: Lag-Llama and AutoGluon, to enhance prediction accuracy in univariate time series forecasting. The process starts with Step a, where both models generate 24-step-ahead forecasts using historical data from selected context windows (1 week, 3 weeks, or 5 weeks). Lag-Llama generates a sample of 100 values for each forecast, and the mean of these values is taken as its point prediction. AutoGluon, on the other hand, provides decile-based (10 percentile - 90 percentile) forecasts, with the 50th percentile value serving as its point prediction. This procedure is applied iteratively as the context window slides forward by one hour. Step b focuses on constructing an ensemble model using three different approaches for each step-ahead forecasts:

\begin{itemize}
    \item Simple Averaging: The forecasts from both models are averaged to create a combined prediction.

    \item Weighted Combinations: Different weight distributions are applied to Lag-Llama and AutoGluon predictions to explore the impact of model dominance in the ensemble (e.g., 40\%-60\%, 25\%-75\%, and 75\%-25\% weightings).

    \item Regression-Based Ensemble: A Linear Regression model is trained on historical forecasted values from Lag-Llama and AutoGluon as input features. The dataset is split into training (80\%) and testing (20\%), preserving time order. The regression model is trained using Time Series Cross-Validation (5 splits) to optimize the weight coefficients for each model's prediction.
\end{itemize}

Finally, in step c, the algorithm evaluates the effectiveness of each individual model and the ensemble approaches by calculating the Mean Squared Error (MSE) between the actual values and the predicted values. This approach enables the selection of an optimal ensemble method that minimizes forecasting errors, leveraging the strengths of both Transformer-based and Statistics-driven time series models.

\subsection{Prediction Intervals for Ensembles}
Let $\mu_{\text{ens}}$ be the ensemble mean and assume component forecast independence. The ensemble variance is
\begin{equation}
\sigma^2_{\text{ens}} = w_1^2 \sigma^2_{\text{Lag}} + w_2^2 \sigma^2_{\text{AG}}.
\end{equation}
We estimate $\sigma_{\text{Lag}}$ from the sample standard deviation of Lag-Llama's $n{=}100$ draws. AutoGluon returns decile forecasts; under a normal approximation we estimate
\begin{equation}
\sigma_{\text{AG}} \approx \frac{P_{90} - P_{10}}{2.5631}.
\end{equation}
A two-sided $95\%$ prediction interval is
\begin{equation}
\mathrm{PI}_{\text{lower}} = \mu_{\text{ens}} - 1.96\,\sigma_{\text{ens}},\qquad
\mathrm{PI}_{\text{upper}} = \mu_{\text{ens}} + 1.96\,\sigma_{\text{ens}}.
\end{equation}

\subsection{Residual Modeling and Iterative Error Feedback}
Residual modeling computes $e_t{=}\hat{y}^{\text{Lag}}_t {-} y_t$ and fits AutoGluon to the residual series; predicted residuals $\hat{e}_t$ are subtracted to form adjusted forecasts $\hat{y}^{\text{adj}}_t {=} \hat{y}^{\text{Lag}}_t {-} \hat{e}_t$. For models that accept exogenous inputs (e.g., TimeGPT), we apply iterative error feedback: compute residuals from an initial pass, append them as an input channel, reforecast, and repeat until validation RMSE converges.

\begin{algorithm}
\caption{Adjusting Lag-LLama Prediction with Statistical Methods}\label{alg:adjust_lagllama_w_error}
\begin{algorithmic}[1]

\Statex \textbf{Input:} Univariate time series data
\Statex \textbf{Output:} Adjusted forecasted values

\Statex \textbf{Step a: Lag-Llama Prediction Process}
    \State Select past \{1 week, 3 weeks, 5 weeks\} as context window. Perform zero-shot forecasting with Lag-Llama to get next hour One-Step-Ahead  sample of 100 forecasts.
    \State Compute the point estimate: $\hat{y}_t = \frac{1}{100} \sum_{i=1}^{100} y_t^{(i)}$
    \State Compute prediction error:
    $e_t = \hat{y}_t - y_t$
    \State Slide the context window by 1 hour and repeat.

\Statex \textbf{Step b: Statistical Approach for Error Prediction}
\State Train a statistical time series model (Autogluon) using error series $e_t$
    \State Select past \{1 week, 3 weeks, 5 weeks\} of error values as input.
    \State Predict the next 1-hour error value: $\hat{e}_t$
    \State Slide the context window by 1 hour and repeat.
    
\Statex \textbf{Step c: Adjust Lag-Llama Predictions using Autogluon}
    \State Retrieve Lag-Llama prediction $\hat{y}_t$
    \State Adjust prediction using estimated error:
    $ \hat{y}^{\text{final}}_t = \hat{y}_t - \hat{e}_t $

\Statex \textbf{Step d: Evaluation}
\State Compute Mean Squared Error (MSE) between:
    \begin{itemize}
        \item Lag-Llama point prediction $\hat{y}_t$ and actual values.
        \item Adjusted prediction $\hat{y}^{\text{final}}_t$ and actual values.
    \end{itemize}

\end{algorithmic}
\end{algorithm}

The Algorithm \ref{alg:adjust_lagllama_w_error}: Adjusting Lag-Llama Prediction with Statistical Methods enhances the accuracy of Lag-Llama forecasts by leveraging statistical modeling to correct prediction errors. The process begins with the Lag-Llama Prediction Process, where past time series data from context windows of 1 week, 3 weeks, or 5 weeks is used to perform zero-shot forecasting. This produces a one-step-ahead forecast sample of 100 values, and the point prediction is computed as the arithmetic mean of this sample. The prediction error, defined as the difference between the predicted and actual values, is also recorded. The context window then slides forward by one hour, repeating the process to capture the evolving error pattern.

To refine predictions, the algorithm employs a Statistical Approach for Error Prediction. Here, AutoGluon, is trained on the historical error series ($e_t$) generated from Lag-Llama predictions. Using the same context window lengths, AutoGluon forecasts the next one-hour error value ($\hat{e}_t$), modeling the systematic biases or fluctuations in Lag-Llama’s output.

In the Adjustment Step, the predicted error ($\hat{e}_t$) is subtracted from the original Lag-Llama forecast ($\hat{y}_t$) to produce the final adjusted prediction ($\hat{y}^{\text{final}}_t$). This correction helps mitigate biases and improve the accuracy of the forecasts by accounting for recurring patterns in the prediction errors.

Finally, in the Evaluation Step, the algorithm assesses performance by computing the Mean Squared Error (MSE) for both the original Lag-Llama predictions and the adjusted forecasts. Comparing these MSE values provides insight into the effectiveness of the statistical adjustment, demonstrating whether the integration of AutoGluon-based error correction enhances predictive accuracy. By systematically learning and correcting errors, this approach refines the robustness of Lag-Llama forecasts, making them more reliable for practical applications.

\section{Experimental Setup}

We use the Belgium Electricity Short-Term Load Forecasting dataset consisting of hourly demand (MW). The series exhibits strong diurnal and weekly seasonality with irregular fluctuations due to weather, markets, and holidays (Appendix~\ref{app:data}). We consider past context windows of 1, 3, and 5 weeks and forecast horizons of 1--24 hours. Data are split chronologically into train/validation/test to avoid leakage and z-normalized for modeling; predictions are back-transformed for evaluation.

Experiments use forecast horizons of 1--24 hours and context lengths of 1, 3, and 5 weeks. The earliest 70\% of observations are for training, the next 10\% for validation, and the most recent 20\% for testing. Foundation models include Lag-Llama (decoder-only) and TimeGPT (encoder--decoder). AutoGluon provides a statistical/ML ensemble baseline and is used for residual modeling. Hyperparameters are tuned on validation. Metrics include MSE for point forecasts, RMSE when assessing iterative feedback, and coverage/width of 95\% PIs.

\section{Results and Discussion}
\paragraph{Bootstrap bagging.} Table \ref{tab:bagging} shows that bagging consistently reduced MSE relative to baseline point forecasts. With a one-week context, MSE fell from 1690 to 1297 (23\% improvement). At three weeks, MSE fell from 1722 to 795 (54\%), and at five weeks from 914 to 486 (47\%).

\begin{table}[h]
\centering
\caption{MSE for Lag-Llama with and without bootstrap bagging (Belgium dataset).}
\label{tab:bagging}
\begin{tabular}{lccc}
\toprule
Context Length & Point Forecast & Bagged Forecast & \% Improvement \\
\midrule
1 week  & 1690 & 1297 & 23\% \\
3 weeks & 1722 & 795  & 54\% \\
5 weeks & 914  & 486  & 47\% \\
\bottomrule
\end{tabular}
\end{table}

\begin{figure}
    \centering
    \includegraphics[width=1\linewidth]{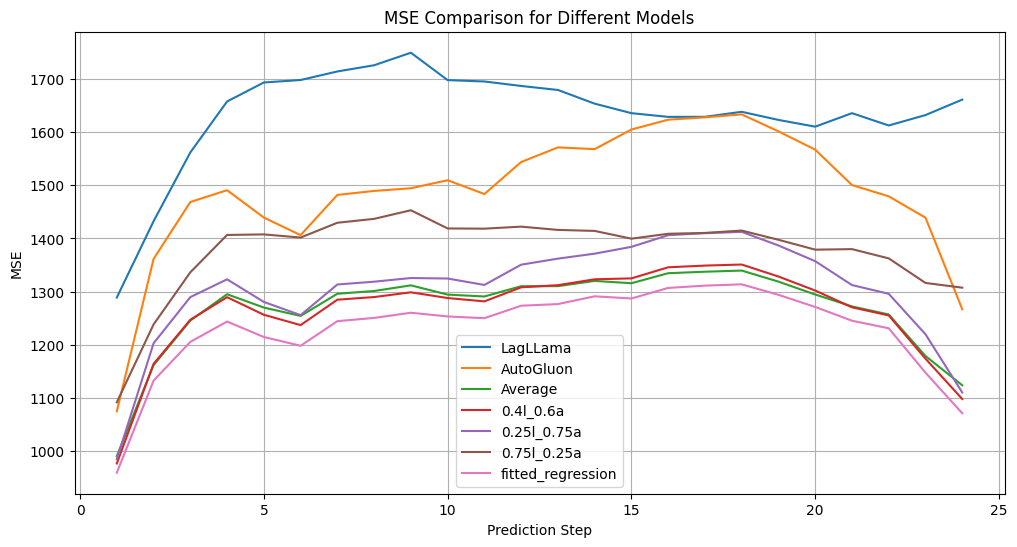}
    \caption[Mse Comparison of Lag-llama, Autogluon and Their Ensemble Models for the next 24 Hours of Forecasting Using 1 Week of Context Length on Belgium Market of Electricity Short-term Load Data.]{MSE comparison of lag-llama, Autogluon and their ensemble models for the next 24 hours of forecasting using 1 week of context length on Belgium Market of Electricity Short-Term Load data.}
    \label{fig:mse_comparison_1w}
\end{figure}

\begin{figure}
    \centering
    \includegraphics[width=1\linewidth]{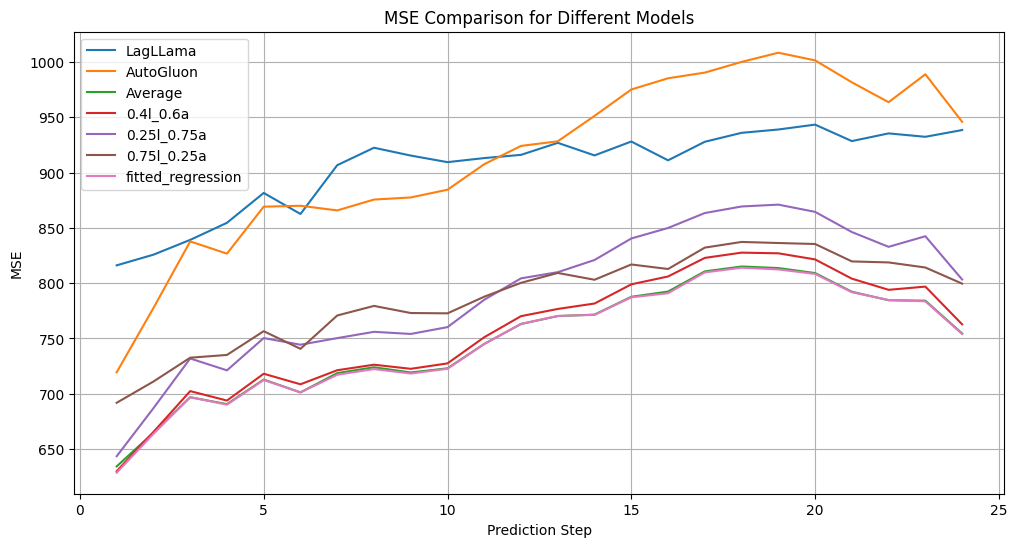}
    \caption[Mse Comparison of Lag-llama, Autogluon and Their Ensemble Models for the next 24 Hours of Forecasting Using 3 Weeks of Context Length on Belgium Market of Electricity Short-term Load Data.]{MSE comparison of lag-llama, Autogluon and their ensemble models for the next 24 hours of forecasting using 3 weeks of context length on Belgium Market of Electricity Short-Term Load data.}
    \label{fig:mse_comparison_3w}
\end{figure}

\begin{figure}
    \centering
    \includegraphics[width=1\linewidth]{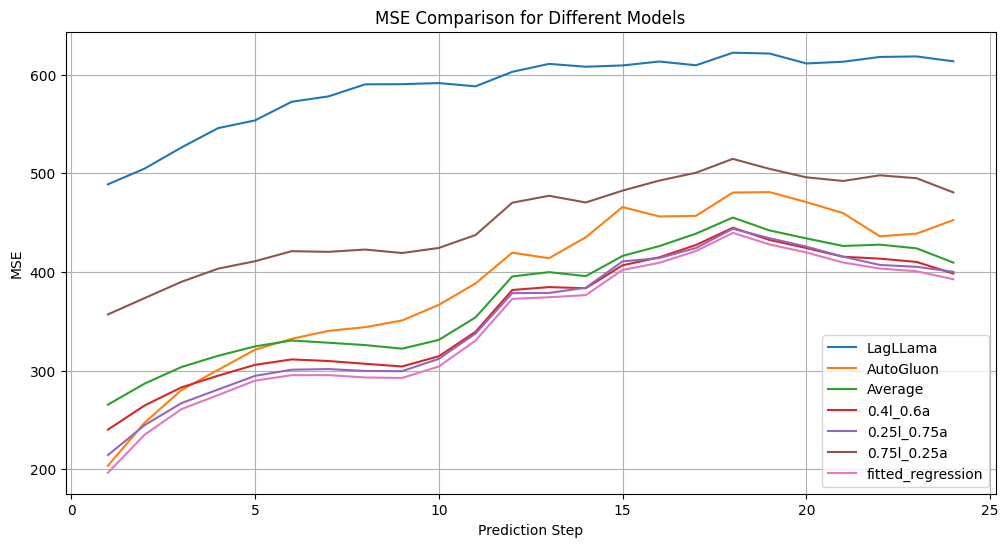}
    \caption[Mse Comparison of Lag-llama, Autogluon and Their Ensemble Models for the next 24 Hours of Forecasting Using 5 Weeks of Context Length on Belgium Market of Electricity Short-term Load Data.]{MSE comparison of lag-llama, Autogluon and their ensemble models for the next 24 hours of forecasting using 5 weeks of context length on Belgium Market of Electricity Short-Term Load data.}
    \label{fig:mse_comparison_5w}
\end{figure}

The results presented in Table \ref{tab:ensemble} are limited to one-step-ahead predictions and do not reflect the performance over longer forecast horizons. To address this, Figures: \ref{fig:mse_comparison_1w}, \ref{fig:mse_comparison_3w}, and \ref{fig:mse_comparison_5w} provide a visual comparison of Mean Squared Error (MSE) for individual models and their ensemble counterparts across step-ahead forecasts from 1 to 24, using context lengths of 1 week, 3 weeks, and 5 weeks, respectively, for the Belgium Electricity Short-Term Load dataset. The x-axis represents the prediction step (1 to 24), while the y-axis indicates the MSE values, offering insights into model performance over extended forecasting horizons.

In Figure \ref{fig:mse_comparison_1w}, Lag-Llama predictions exhibit the highest MSE, followed by AutoGluon, then the simple average, with the fitted regression ensemble model showing the lowest MSE across all steps. As the forecast horizon increases, the MSE for all models tends to rise, indicating that model performance generally degrades with longer forecasting horizons. However, an interesting observation is that after step 17, the models appear to show an improvement in performance, contrary to expectations. This unexpected trend could be attributed to the shorter context length (1 week), which may limit the model’s ability to accurately capture long-term patterns and trends, leading to inconsistent performance beyond a certain prediction step.

In Figure \ref{fig:mse_comparison_3w}, the MSE values of Lag-Llama and AutoGluon intertwine, with Lag-Llama showing higher MSE up to step 11, after which AutoGluon exhibits higher errors. The ensemble models consistently outperform the individual models, with the fitted regression ensemble demonstrating the best overall performance across the 24-step forecast horizon. As the forecast horizon increases, the MSE for all models tends to rise, indicating that model performance generally degrades with longer forecasting horizons. 

In Figure \ref{fig:mse_comparison_5w}, Lag-Llama exhibits the highest MSE values, followed by AutoGluon, with the ensemble models performing better overall. The fitted regression ensemble model achieves the lowest MSE, indicating the most accurate predictions. Additionally, as the forecast horizon increases, the MSE consistently rises for all models, highlighting the common trend of decreasing model performance with longer-term forecasts. 

\paragraph{Regression ensembles.} Since the fitted ensemble regression model has demonstrated the best performance based on MSE comparisons across all the step-ahead prediction and context lengths, we further analyze its coefficients to understand the contribution of Lag-Llama and AutoGluon in generating 24-step-ahead forecasts. It is found that linear stacking of Lag-Llama and AutoGluon achieved the lowest MSE across contexts. For a one-week context, as shown in Table \ref{tab:ensemble}, the ensemble obtained 959 versus 1288 (Lag-Llama) and 1074 (AutoGluon). For three weeks, the ensemble achieved 628 versus 816 and 719; for five weeks, it reached 196 versus 488 and 203.

\begin{table}[h]
\centering
\caption{MSE for individual models and regression ensemble (1-step ahead).}
\label{tab:ensemble}
\begin{tabular}{lcccc}
\toprule
Context Length & Lag-Llama & AutoGluon & Avg. Ensemble & Regression Ensemble \\
\midrule
1 week  & 1288 & 1074 & 990 & 959 \\
3 weeks &  816 &  719 & 634 & 628 \\
5 weeks &  488 &  203 & 265 & 196 \\
\bottomrule
\end{tabular}
\end{table}

\begin{figure}
    \centering
    \includegraphics[width=1\linewidth]{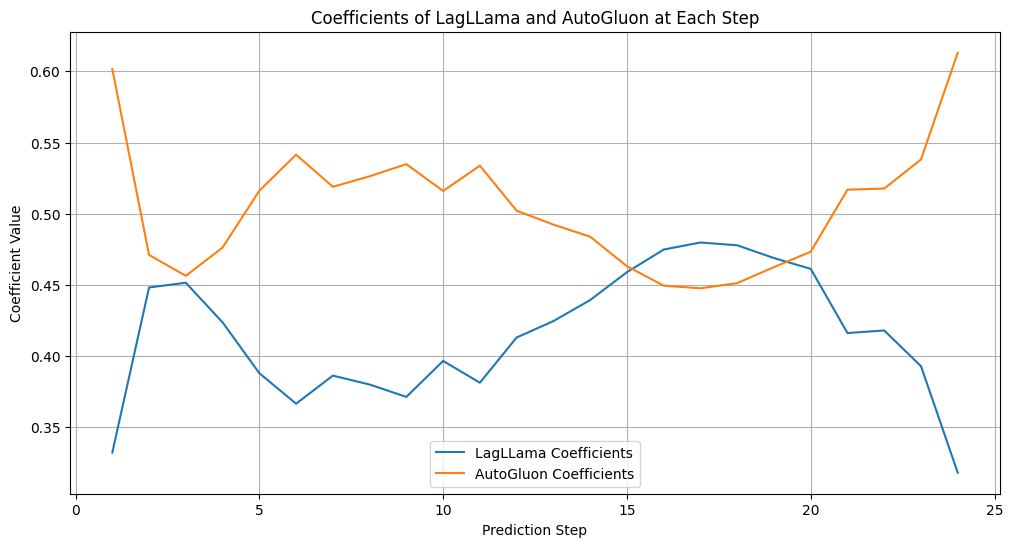}
    \caption[Weights Corresponding to Lag-llama and Autgoluon for the Fitted Regression Model for the next 24 Hours of Forecasting Using 1 Week of Context Length on Belgium Market of Electricity Short-term Load Data.]{Weights corresponding to lag-llama and Autgoluon for the fitted regression model for the next 24 hours of forecasting using 1 week of context length on Belgium Market of Electricity Short-Term Load data.}
    \label{fig:coefficients_1W}
\end{figure}

In the Figure \ref{fig:coefficients_1W}, The Lag-Llama coefficients begin at 0.332 for step 1 and peak at 0.479 around step 17, before declining to 0.318 by step 24. The AutoGluon coefficients start out higher at 0.602, fluctuate a little, peak at step 1, and then progressively drop to 0.613 at step 24. AutoGluon is a major contributor towards fitted regression ensemble model than Lag-Llama for the initial forecast horizon (up to step 12).

\begin{figure}
    \centering
    \includegraphics[width=1\linewidth]{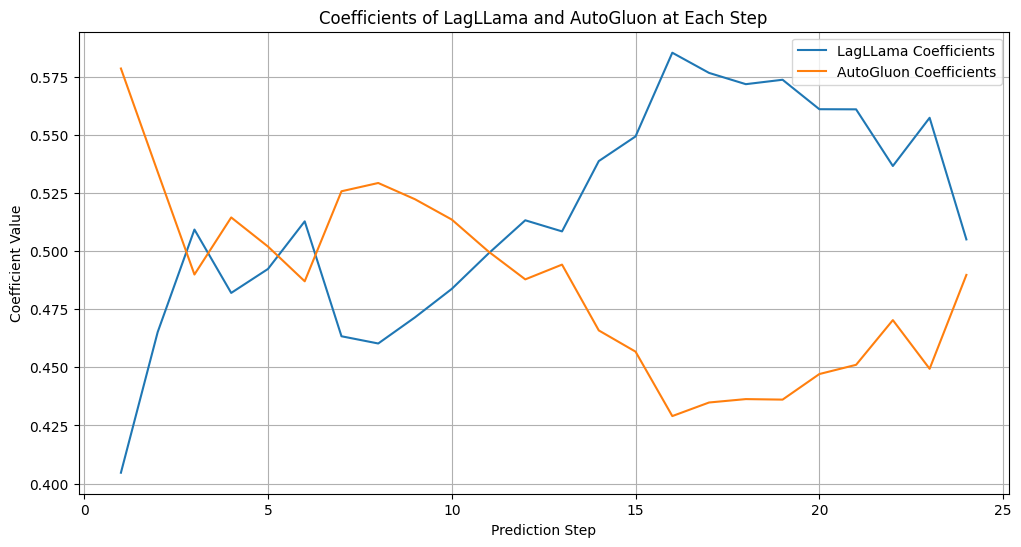}
    \caption[Weights Corresponding to Lag-llama and Autgoluon for the Fitted Regression Model for the next 24 Hours of Forecasting Using 3 Weeks of Context Length on Belgium Market of Electricity Short-term Load Data.]{Weights corresponding to lag-llama and Autgoluon for the fitted regression model for the next 24 hours of forecasting using 3 weeks of context length on Belgium Market of Electricity Short-Term Load data.}
    \label{fig:coefficients_3W}
\end{figure}

In the Figure \ref{fig:coefficients_3W}, starting at 0.405, the Lag-Llama coefficients progressively increase to 0.585 at step 16, and by step 24, they have slightly decreased to 0.505. Starting at 0.579, the AutoGluon coefficients fluctuate and peak at step 4 (0.515). At step 24, they exhibit a downward trend towards 0.490. Lag-Llama exhibits greater contribution towards fitted regression ensemble model prediction over time than the 1-week model, peaking at step 16.

\begin{figure}
    \centering
    \includegraphics[width=1\linewidth]{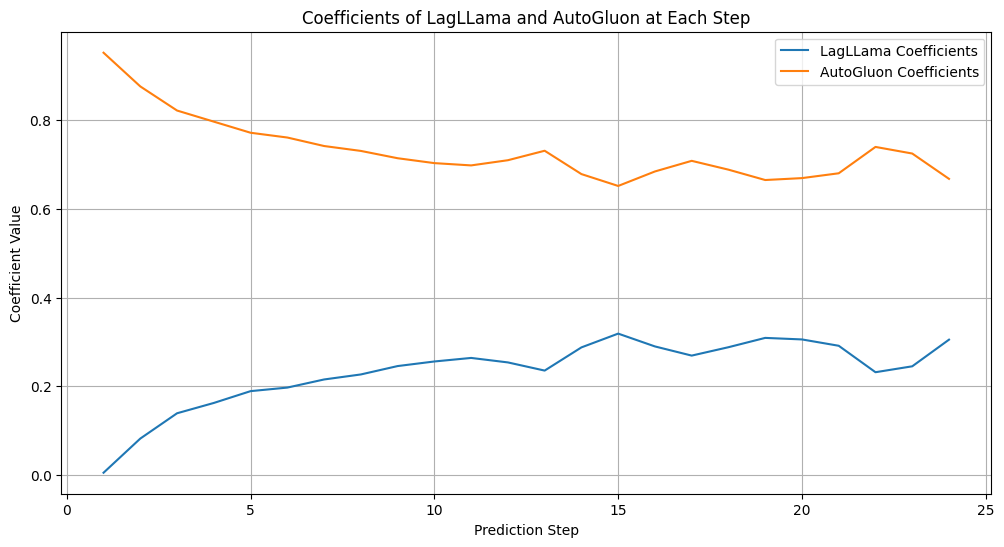}
    \caption[Weights Corresponding to Lag-llama and Autgoluon for the Fitted Regression Model for the next 24 Hours of Forecasting Using 5 Weeks of Context Length on Belgium Market of Electricity Short-term Load Data.]{Weights corresponding to lag-llama and Autgoluon for the fitted regression model for the next 24 hours of forecasting using 5 weeks of context length on Belgium Market of Electricity Short-Term Load data.}
    \label{fig:coefficients_5W}
\end{figure}

In the Figure \ref{fig:coefficients_5W}, starting very low at 0.005, the Lag-Llama coefficients gradually increase until reaching a peak of 0.319 at step 15, after which they fluctuate slightly before reaching 0.305. The AutoGluon coefficients begin at 0.952 and end at 0.668, remaining relatively high but exhibiting a decreasing trend over steps. While AutoGluon maintains a strong predictive power across steps but slightly decreases, the Lag-Llama model shows an initially poor predictive power that steadily improves when compared to the shorter context lengths.

In general, the stability and magnitude of Lag-Llama coefficients increase with the length of the context. The 3-week model offers a smoother and more steady trend, whereas the 1-week model varies more. The 5-week model is quite weak at first, but it gradually gets better over time. AutoGluon coefficients always start out higher than Lag-Llama, but as context length increases, a steady decrease is seen. While the 5-week model gradually loses coefficients, the 1-week and 3-week models exhibit a fair amount of stability. Therefore, increasing the context length makes Lag-Llama more stable but gradually reduces AutoGluon's efficacy. AutoGluon seems to work best with shorter context lengths, whereas Lag-Llama benefits from a longer historical window.

\paragraph{Prediction intervals.} Ensemble PIs achieved near-nominal coverage with widths decreasing as context length increased; moving from one-week to three-weeks reduced average PI width by roughly 20\% without loss of coverage.

\begin{figure}
    \centering
    \includegraphics[width=1\linewidth]{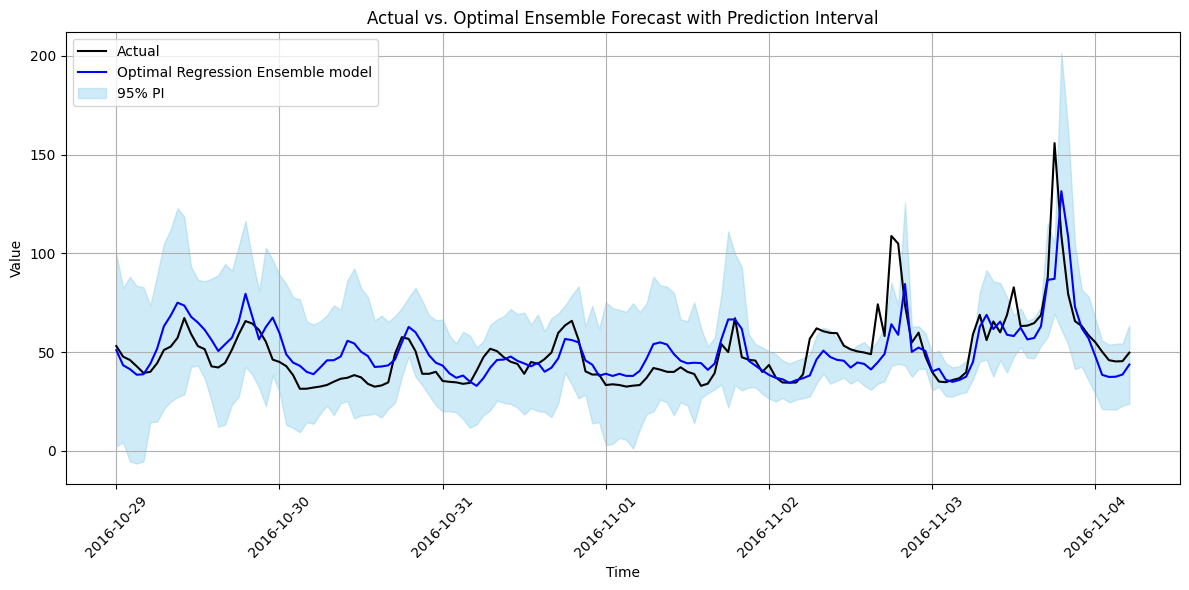}
    \caption[95\% Prediction Interval for the Fitted Regression Model for Step 1 Ahead Forecast for 1 Week of Context Length on Belgium Market of Electricity Short-term Load Data.]{95\% Prediction Interval for the fitted regression model for Step 1 ahead Forecast for 1 week of context length on Belgium Market of Electricity Short-Term Load data.}
    \label{fig:PI_1W}
\end{figure}

\begin{figure}
    \centering
    \includegraphics[width=1\linewidth]{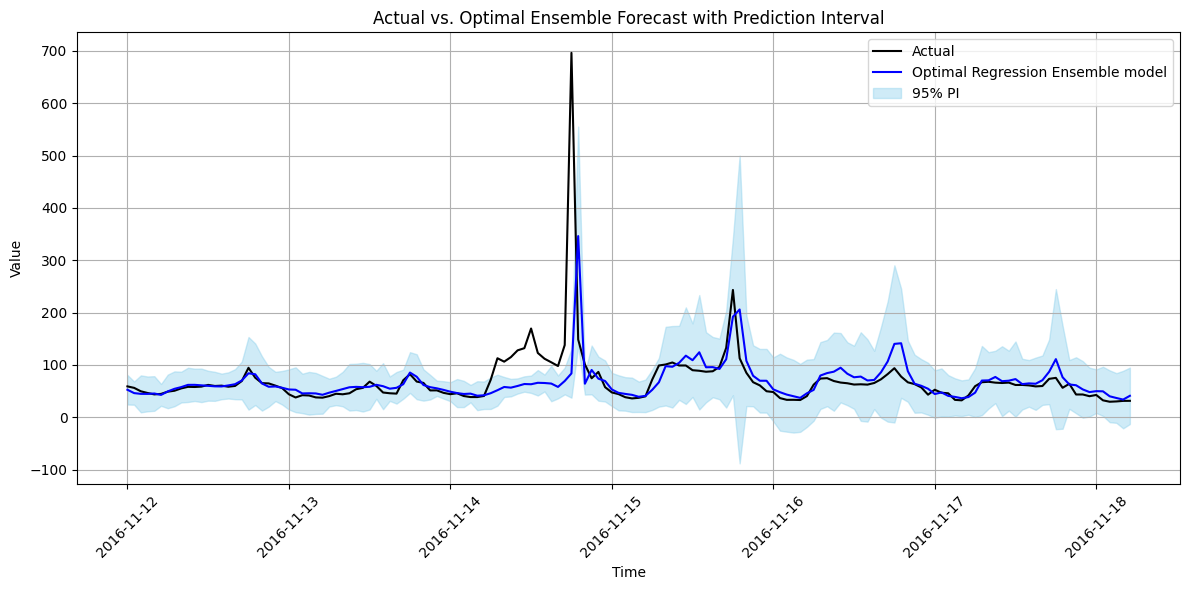}
    \caption[95\% Prediction Interval for the Fitted Regression Model for Step 1 Ahead Forecast for 3 Weeks of Context Length on Belgium Market of Electricity Short-term Load Data.]{95\% Prediction Interval for the fitted regression model for Step 1 ahead Forecast for 3 weeks of context length on Belgium Market of Electricity Short-Term Load data.}
    \label{fig:PI_3W}
\end{figure}

\begin{figure}
    \centering
    \includegraphics[width=1\linewidth]{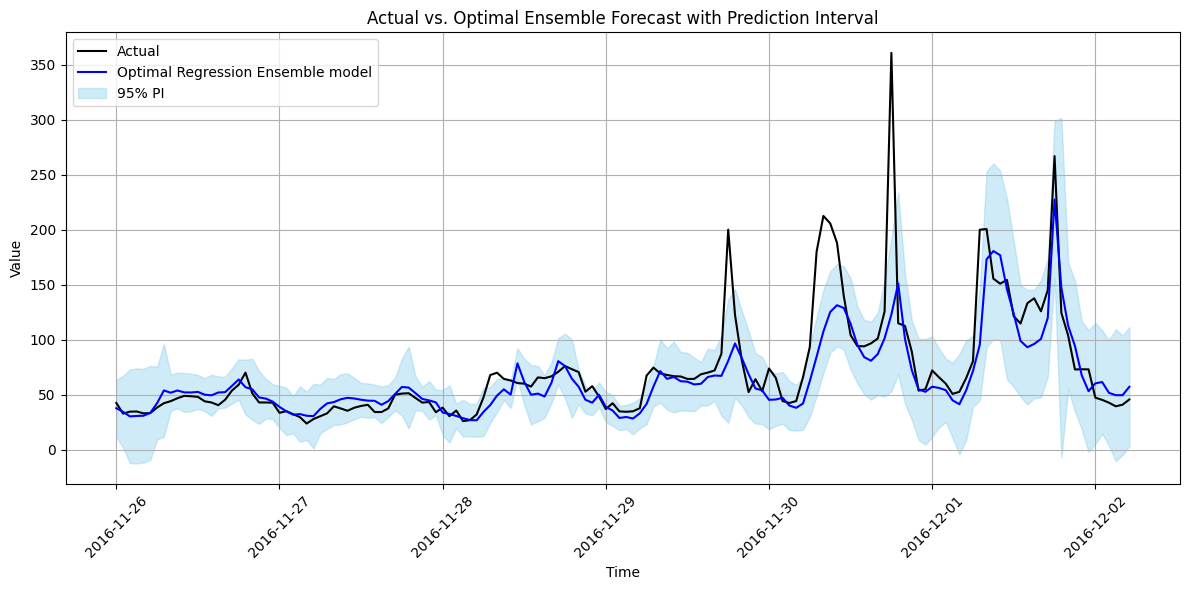}
    \caption[95\% Prediction Interval for the Fitted Regression Model for Step 1 Ahead Forecast for 5 Weeks of Context Length on Belgium Market of Electricity Short-term Load Data.]{95\% Prediction Interval for the fitted regression model for Step 1 ahead Forecast for 5 weeks of context length on Belgium Market of Electricity Short-Term Load data.}
    \label{fig:PI_5W}
\end{figure}

Figures \ref{fig:PI_1W}, \ref{fig:PI_3W}, and \ref{fig:PI_5W} provide a comprehensive view of the predictive accuracy and confidence of the model in different forecasting scenarios. As evident, the prediction interval of the model with context length 1 week $>$ 3 week $>$ 5 week. This suggests that while longer context lengths produce more stable and confident forecasts with narrower intervals, shorter context lengths lead to higher uncertainty, as evidenced by the wider prediction intervals. According to the observed trend, adding a more extensive historical context improves the model's ability to capture long-term dependencies, which lowers prediction variability and increases forecast reliability overall.

\paragraph{Residual modeling.} Training AutoGluon on residuals led to modest gains for short contexts (1\% at one week) and substantial gains for mid-range contexts (67\% at three weeks: 180 to 59 MSE). Five-week evaluation was infeasible due to limited residual history.

\begin{table}[h]
\centering
\caption{MSE for Lag-Llama with and without residual correction.}
\label{tab:residual}
\begin{tabular}{lccc}
\toprule
Context Length & Lag-Llama & Residual-Corrected & \% Improvement \\
\midrule
1 week  & 1413 & 1397 & 1\% \\
3 weeks &  180 &   59 & 67\% \\
5 weeks &  --  &  --  & -- \\
\bottomrule
\end{tabular}
\end{table}

\paragraph{Iterative error feedback.} Providing residuals as exogenous inputs to TimeGPT produced moderate RMSE gains in dynamic settings with drifting residual patterns. While smaller than ensembling gains, this adds adaptability for online scenarios with evolving regimes.

\subsection{Practical Notes}
We observed that bagging benefits increase with context length, consistent with reduced sampling noise in the forecast distribution. Ensemble weights adapt to horizon and context: AutoGluon dominates short contexts and near horizons, whereas Lag-Llama gains influence with longer contexts. PI calibration degrades if component-forecast independence is strongly violated; a pragmatic fix is to inflate $\sigma_{\text{ens}}$ by a small factor estimated on validation.

\section{Conclusion and Future Work}
We presented a unified framework for augmenting Transformer-based foundation models with statistical bagging, stacking, prediction intervals, and residual boosting. On an operational electricity load forecasting task, the hybrids consistently improved accuracy and reliability: regression ensembles achieved the best overall MSE; bagging reduced variance, especially with longer context; and residual modeling corrected mid-horizon bias. Ensemble PIs were well calibrated and narrowed with increased context. Future work includes testing on multivariate and multimodal datasets, online learning variants for streaming updates, interpretability tools for hybrid pipelines, and incorporation of domain constraints into ensemble weighting and residual correction.

\paragraph{Limitations.} Our independence assumption for ensemble PI construction is an approximation; dependencies between components may require copula-based or bootstrap-based PI calibration. Computational costs for bagging grow with the number of bootstrap draws; adaptive subsampling may preserve gains at lower cost.

\appendix
\section*{Appendices}
\section{Electricity Short Duration Load Forecasting Dataset}\label{app:data}
We use hourly electricity load from Belgium, along with exploratory analysis showing strong daily/weekly seasonality and occasional holiday effects. The same data source provides Germany, France, and Nord Pool markets; we restrict attention to Belgium for clarity. Data were cleaned for missingness via simple linear interpolation when needed and standardized before modeling. (See Modi's thesis \citep{modi2025foundation} Appendix A for plots and additional details.)

\section{Regression Coefficients from Stacking}
Table \ref{tab:coeffs} lists horizon-specific weights $(w_1,w_2)$ from the fitted regression stacks across 1--24 step-ahead forecasts and each context length. In practice, weights shift towards AutoGluon for shorter contexts and towards Lag-Llama for longer contexts, consistent with their complementary inductive biases.

\begin{table}[h]
    \centering
    \caption[Optimal Regression Models' Co-efficients for Lag-llama and Autogluon Predictions Trained for All 24 Step Ahead Predictions Across Context Length of 1 Week, 3 Weeks and 5 Weeks on Belgium Market of Electricity Short-term Load Data]{Optimal Regression Models' Co-efficients for Lag-llama and AutoGluon predictions trained for all 24 step ahead predictions across context length of 1 week, 3 weeks and 5 weeks on Belgium market of Electricity Short-Term Load data}
    \vspace{6pt}
    \begin{tabular}{ccccccc}
    \hline
    &\multicolumn{2}{c}{1 week}&\multicolumn{2}{c}{3 weeks}&\multicolumn{2}{c}{5 weeks}\\
    Step& Lag-Llama& AutoGluon& Lag-Llama& AutoGluon& Lag-Llama& AutoGluon\\ \hline
    1& 0.332125& 0.601680& 0.404662& 0.578614& 0.005344& 0.952155\\
    2& 0.448116& 0.471027& 0.465180& 0.534136& 0.082346& 0.876162\\
    3& 0.451454& 0.456282& 0.509311& 0.489959& 0.139380& 0.821817\\
    4& 0.423607& 0.476178& 0.482038& 0.514523& 0.162635& 0.796770\\
    5& 0.387868& 0.516094& 0.492347& 0.502000& 0.189260& 0.771744\\
    6& 0.366428& 0.541459& 0.512842& 0.487044& 0.197277& 0.760975\\
    7& 0.386173& 0.518859& 0.463390& 0.525784& 0.215566& 0.741829\\
    8& 0.379898& 0.526257& 0.460277& 0.529347& 0.226889& 0.730793\\
    9& 0.371221& 0.534776& 0.471524& 0.522355& 0.245775& 0.714312\\
    10& 0.396490& 0.515927& 0.483771& 0.513617& 0.256128& 0.703277\\
    11& 0.381177& 0.533720& 0.499019& 0.499841& 0.264117& 0.698130\\
    12& 0.412973& 0.502074& 0.513298& 0.487874& 0.254007& 0.709828\\
    13& 0.424469& 0.492320& 0.508519& 0.494223& 0.235598& 0.731183\\
    14& 0.439280& 0.483804& 0.538777& 0.465902& 0.287869& 0.678569\\
    15& 0.458843& 0.463032& 0.549388& 0.456768& 0.318937& 0.651808\\
    16& 0.474764& 0.449383& 0.585411& 0.429040& 0.290040& 0.684475\\
    17& 0.479718& 0.447538& 0.576720& 0.434884& 0.269370& 0.708433\\
    18& 0.477774& 0.451161& 0.571875& 0.436345& 0.288281& 0.688541\\
    19& 0.468783& 0.462445& 0.573778& 0.436123& 0.309296& 0.665185\\
    20& 0.461191& 0.473239& 0.561101& 0.447128& 0.305814& 0.669495\\
    21& 0.416090& 0.516813& 0.561045& 0.451100& 0.291534& 0.680372\\
    22& 0.417893& 0.517644& 0.536663& 0.470341& 0.231913& 0.739794\\
    23& 0.392781& 0.538055& 0.557414& 0.449382& 0.245241& 0.724904\\
    24& 0.317927& 0.613094& 0.505071& 0.489788& 0.305317& 0.667808\\\hline
    \end{tabular}
    \label{tab:coeffs}
\end{table}

\section{Code Availability}
The source code and experimental notebooks supporting the results of this thesis are publicly available at:

\url{https://github.com/DhruvModi1605/enhancing\_foundation\_time\_series\_model\_prediction}

\bibliographystyle{plainnat}
\bibliography{references3}
\end{document}